\documentclass[runningheads]{llncs}
\usepackage{graphicx}
\usepackage{amsmath}
\usepackage{amsfonts}
\usepackage{amssymb}
\usepackage{color}
\usepackage{tikz}
\usetikzlibrary{arrows}

\newcommand{\shrink}[1]{{}}%

\def\eql(#1,#2){{#1\!\!=\!#2}}
\def\neql(#1,#2){{#1\!\!\not = \!#2}}

\def\n#1{\bar{#1}}

\def\t{{\tau}}
\def\r{{\rho}}
\def\c{{\alpha}}

\def\l{{\ell}}

\begin{document}

\title{On Symbolically Encoding the\\ Behavior of Random Forests}

\author{Arthur Choi\inst{1} \and
Andy Shih\inst{2} \and
Anchal Goyanka\inst{1} \and
Adnan Darwiche\inst{1}}
\authorrunning{A. Choi et al.}
% First names are abbreviated in the running head.
% If there are more than two authors, 'et al.' is used.
%
\institute{Computer Science Department, UCLA\\
\email{$\{$aychoi, anchal, darwiche$\}$@cs.ucla.edu} \and
Computer Science Department, Stanford University\\
\email{andyshih@cs.stanford.edu}}

\maketitle

\begin{abstract}
Recent work has shown that the input-output behavior of some machine learning systems can be captured symbolically
using Boolean expressions or tractable Boolean circuits, which facilitates reasoning about the behavior of these
systems. While most of the focus has been on systems with Boolean inputs and outputs, we address
systems with discrete inputs and outputs, including ones with discretized continuous variables as in systems 
based on decision trees.
We also focus on the suitability of encodings for computing prime implicants, which
have recently played a central role in explaining the decisions of machine learning systems. We show
some key distinctions with encodings for satisfiability, and propose an encoding that is sound and 
complete for the given task. 

\end{abstract}

\section{Introduction}

Recent work has shown that the input-output behavior of some machine learning systems can be captured symbolically
using Boolean expressions or tractable Boolean circuits \cite{KatzBDJK17,Leofante18,NarodytskaKRSW18,ShihCD18,IgnatievNM19a,IgnatievNM19b,SDC19,shi2020tractable}. These encodings facilitate the reasoning about the behavior of these
systems, including the explanation of their decisions, the quantification of their robustness and the verification of their
properties. Most of the focus has been on systems with Boolean inputs and outputs, with some extensions to discrete inputs and outputs, including discretizations of continuous variables as in systems based on
decision trees; see, e.g., \cite{DBLP:conf/cp/BessiereHO09,NarodytskaIPM18,JoaoApp}. This paper is concerned with the latter case of discrete/continuous 
systems but those that are encoded using Boolean variables, with the aim of utilizing the vast machinery available for reasoning with Boolean logic.
Most prior studies of Boolean encodings have focused on the tasks of satisfiability
and model counting \cite{DBLP:conf/ijcai/Kleer89,DBLP:conf/cp/Walsh00,DBLP:conf/cp/BessiereHO09}. In this paper, we focus instead on prime implicants which
have recently played a central role in explaining the decisions of machine learning systems \cite{ShihCD18,Renooij18,IgnatievNM19a,IgnatievNM19b,JoaoApp,DarwicheHirth20a}; cf. \cite{anchors:aaai18}. 
We first highlight how the prime implicants of a multi-valued expression are not immediately obtainable as prime implicants of a corresponding Boolean encoding.  We reveal how to compute these prime implicants, by computing them instead on a Boolean expression derived from the encoding.  
Our study is conducted in the context of encoding the behavior of random forests using majority voting, but our results apply more broadly.

This paper is structured as follows.
We introduce the task in Section~\ref{sec:behaviors} as well as review related work.  We discuss in Section~\ref{sec:explanation} the problem of explaining the decisions of machine learning systems whose continuous features can be discretized into intervals. We follow in Section~\ref{sec:encodings} by a discussion on encoding the input-output behavior 
of such systems, where we analyze three encodings from the viewpoint of computing
explanations for decisions. We show that one of these encodings is suitable for this purpose, if employed
carefully, while proving its soundness and completeness for the given task.
We finally close in Section~\ref{sec:conclusion}. 

\section{Boolean, Discrete and Continuous Behaviors}
\label{sec:behaviors}

The simplest behaviors to encode are for systems with Boolean inputs and outputs. Consider a neural network
whose inputs are Boolean and that has only step activation functions. Each neuron in this network computes a
Boolean function and therefore each output of the network also computes a Boolean function. 
The input-output behavior of such networks can be immediately represented using Boolean expressions, or 
Boolean circuits as proposed in~\cite{ChoiShiShihDarwiche18,shi2020tractable}.

Suppose now that the inputs to a machine learning system are discrete variables, say, 
variable \(A\) with values \(1,2,3\), variable \(B\) with values \(r, b, g\) and  variable \(C\) with values \(l, m, h\).
One can define a multi-valued propositional logic to capture the behavior of such a system.
The atomic expressions in this case will be of the form \(\eql(V,v)\), indicating that discrete variable \(V\)
has the value \(v\). We can then construct more complex expressions using Boolean connectives. 
An example expression in this logic would be \((\eql(B,r) \vee \eql(B,b)) \implies (\eql(A,2) \wedge \neg \eql(C,h))\).

\begin{figure}[t]
\centering
\raisebox{-.5\height}{\includegraphics[width=.4\linewidth]{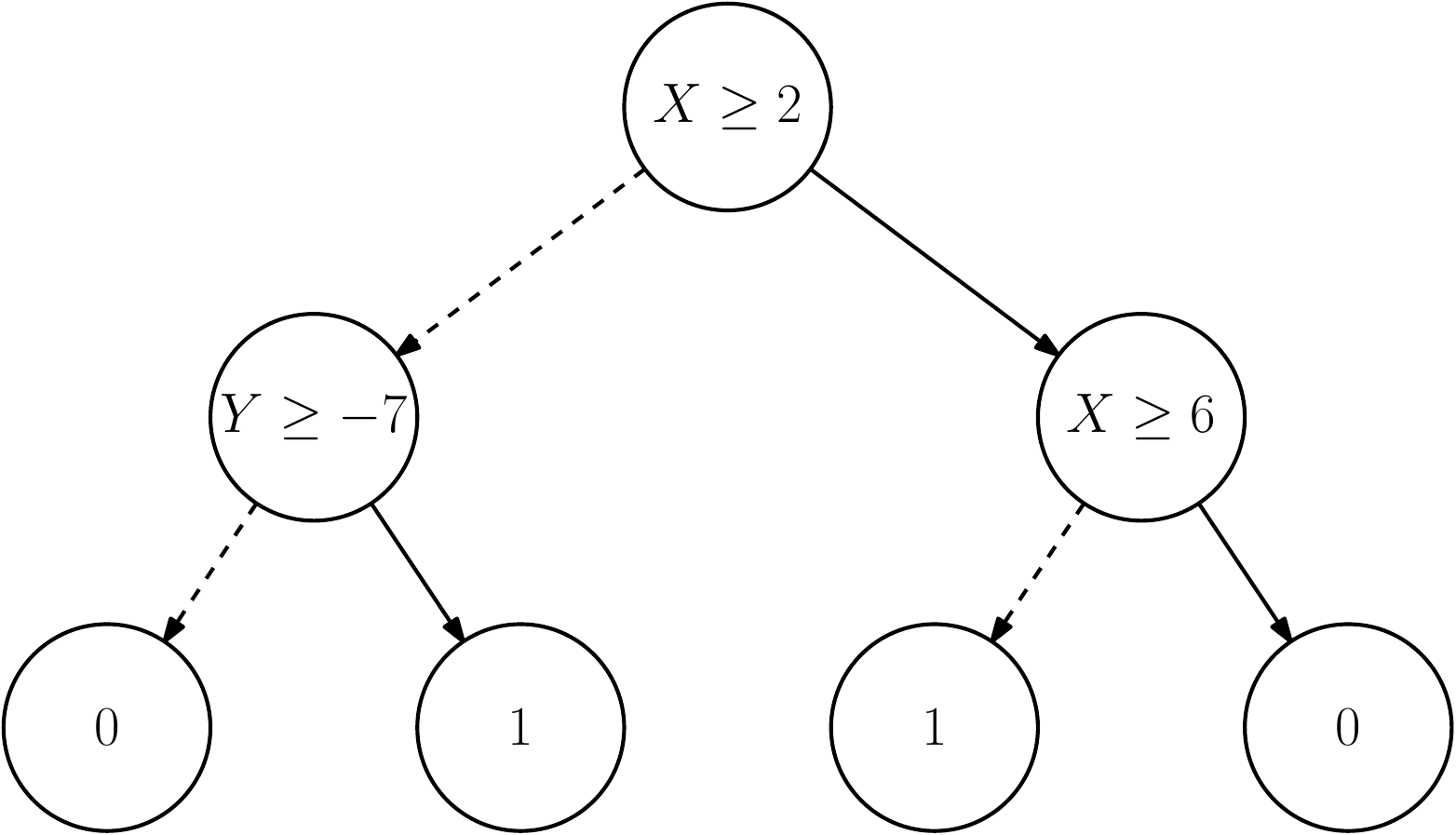}}
\quad
\begin{tabular}{c|c} 
  value & interval \\ \hline
  \(x_1\) & \((-\infty, 2)\) \\
  \(x_2\) & \([2, 6)\) \\
  \(x_3\) & \([6,+\infty)\) \\ \hline
  \(y_1\) & \((-\infty, -7)\) \\
  \(y_2\) & \([-7,+\infty)\) \\
\end{tabular}
\quad
\begin{tabular}{cc|c}
  \(X\)    & \(Y\)    & \(f(X,Y)\)    \\ \hline
  \(x_1\)  & \(y_1\)  & 0 \\
  \(x_1\)  & \(y_2\)  & 1 \\
  \(x_2\)  & \(y_1\)  & 1 \\
  \(x_2\)  & \(y_2\)  & 1 \\
  \(x_3\)  & \(y_1\)  & 0 \\
  \(x_3\)  & \(y_2\)  & 0 \\
\end{tabular}
\caption {(Left) A decision tree of continuous variables \(X\) and \(Y\), where a solid branch means the test is true, and a dashed branch means false.  (Center) A discretization of \(X\) and \(Y\) into intervals.  (Right) The discrete function represented by the decision tree. \label{fig:dt}}
\end{figure}

Some systems may have continuous variables as inputs, which get discretized during the learning process
as is the case with systems based on decision trees. Consider for example the decision tree
in Figure~\ref{fig:dt} (left) over continuous variables \(X\) and \(Y\). The algorithm that learned this
tree discretized its variables as follows:
%\begin{eqnarray*}
%X : & & (-\infty, 2),\:\: [2, 6), \:\: [6,+\infty) \\
%Y : & & (-\infty, -7), \:\: [-7,+\infty)
%\end{eqnarray*}
\(X\) to intervals \((-\infty, 2), [2, 6), [6,+\infty)\) and 
\(Y\) to intervals \((-\infty, -7), [-7,+\infty).\)

We can now think of variable \(X\) as a discrete variable with three values \(x_1, x_2, x_3\), each corresponding to one of the 
intervals as shown in Figure~\ref{fig:dt} (center).
Variable \(Y\) is binary in this case, with each value corresponding to one of the two intervals. According to this
decision tree, the infinite number of input values for variables \(X\) and \(Y\) can be grouped into six equivalence classes as shown
in Figure~\ref{fig:dt} (right). Hence, the input-output behavior of this decision tree can be captured using the multi-valued propositional
expression \(f(X,Y) = (\eql(X,x_1)\wedge\eql(Y,y_2)) \vee \eql(X,x_2)\), even though we have continuous variables.

Our goal is therefore to encode multi-valued expressions using Boolean expressions as we aim to exploit the vast machinery 
currently available for reasoning with propositional logic. This includes SAT-based and knowledge compilation tools, 
which have been used extensively recently to reason about the behavior of machine learning systems
\cite{KatzBDJK17,Leofante18,NarodytskaKRSW18,ShihCD18,IgnatievNM19a,IgnatievNM19b,SDC19,shi2020tractable}.

Encoding multi-valued expressions using Boolean expressions has been of interest for a very long time and
several methods have been proposed for this purpose; see, e.g.,~\cite{DBLP:conf/ijcai/Kleer89,DBLP:conf/cp/Walsh00,DBLP:conf/cp/BessiereHO09}.
In some cases, different encodings have been compared in terms of the efficacy of applied SAT-based tools;
see, e.g.,~\cite{DBLP:conf/cp/Walsh00}. In this paper, we consider another dimension for evaluating encodings,
which is based on their suitability for computing prime implicants. This is motivated by the fundamental role that implicants 
have been playing recently in explaining the decisions of machine learning systems~\cite{ShihCD18,Renooij18,IgnatievNM19a,IgnatievNM19b,JoaoApp,DarwicheHirth20a}.

The previous works use the notion of a {\em PI-explanation} when explaining the decision of a classifier on an instance. 
A PI-explanation, introduced in~\cite{ShihCD18}, is a minimal set of instance characteristics that are sufficient to
trigger the decision. That is, if these characteristics are fixed, other instance characteristics can be
changed freely without changing the decision.
In an image, for example, a PI-explanation corresponds to a minimal set of pixels that guarantees the stability of a decision 
against any perturbation of the remaining pixels.\footnote{A PI-explanation can be viewed as a (minimally)  {\em sufficient
reason} for the decision~\cite{DarwicheHirth20a}.}

PI-explanations are based on {\em prime implicants} of Boolean functions, which have been
studied extensively in the literature~\cite{BooleanFunctions,quine1,mccluskey,quine2}.
Consider the following Boolean function over variables \(A\), \(B\) and \(C\): \(f = (A + \overline{C})(B + C)(A + B).\)
A prime implicant of the function is a minimal setting of its variables that causes the
function to trigger. This function has three prime implicants: \(AB\), \(AC\) and \(B\overline{C}\).
Consider now the instance \(AB\overline{C}\) leading to a positive decision \(f(AB\overline{C})=1\). 
The PI-explanations for this decision are the prime implicants of function \(f\) that are 
compatible with the instance: \(AB\) and \(B\overline{C}\).
Explaining negative decisions requires working with the function's complement \(\overline{f}\).
Consider instance \(\overline{A}BC\), which sets the function \(f\) to~\(0\). The complement \(\overline{f}\)
has three prime implicants \(\overline{A}C\), \(\overline{B}\,\overline{C}\) and \(\overline{A}\,\overline{B}\).
Only one of these is compatible with the instance, \(\overline{A}C\), so it is the
only PI-explanation for the decision on this instance.\footnote{The popular 
Anchor system~\cite{ANCHOR} can be viewed as computing approximations of PI-explanations. 
The quality of these approximations has been evaluated on some datasets and corresponding 
classifiers in~\cite{JoaoApp}, where an approximation is called {\em optimistic} if it is a strict subset 
of a PI-explanation and  {\em pessimistic} if it is a strict superset of a PI-explanation.
Anchor computes approximate explanations without having to abstract the machine learning system behavior into a symbolic representation.}

When considering the encoding of multi-valued expressions using Boolean ones, we will be focusing on whether 
the prime implicants of multi-valued expressions can be soundly and completely obtained from the prime implicants of the 
corresponding Boolean expressions. This is motivated by the desire to exploit existing algorithms and tools for computing prime implicants of
Boolean expressions (one may also consider developing a new set of algorithms and tools for
operating directly on multi-valued expressions).

Before we propose and evaluate some encodings, we need to first define the notion of a prime implicant for multi-valued
expressions and then examine explanations from that perspective. This is needed to settle the semantics of explanations in a
multi-valued setting, which will then form the basis for deciding whether a particular encoding is satisfactory from the viewpoint of computing
explanations. As the following discussion will reveal, the multi-valued
setting leads to some new considerations that are preempted in a Boolean setting.

\section{Explaining Decisions in a Multi-Valued Setting}
\label{sec:explanation}

Consider again the decision tree in Figure~\ref{fig:dt} whose behavior is captured by the multi-valued expression
\((\eql(X,x_1)\wedge\eql(Y,y_2)) \vee \eql(X,x_2)\) as discussed earlier. Consider also the positive instance
\(\eql(X,3) \wedge \eql(Y,12)\), which can be represented using the
multi-valued expression \(\alpha: \eql(X,x_2) \wedge \eql(Y,y_2)\) as shown in Figure~\ref{fig:dt}.

Instance \(\alpha\) has two characteristics \(\eql(X,x_2)\) and \(\eql(Y,y_2)\), yet one of them \(\eql(X,x_2)\) is sufficient to trigger the
positive decision. Hence, one explanation for the decision is that variable \(X\) takes a value
in the interval \([2,6)\), which justifies \(\eql(X,x_2)\) as a PI-explanation of this positive decision. In fact, if we stick to
the literal definition of a PI-explanation from the Boolean setting, then this would be the only PI-explanation
since \(\eql(Y,y_2)\) is the only characteristic that can be dropped from the instance while guaranteeing that the
decision will stick.

Looking closer, this decision would also stick if the value of \(X\) were contained in the larger
interval \((-\infty,6)\) as long as characteristic \(\eql(Y,y_2)\) is maintained. The interval \((-\infty,6)\) corresponds to 
\((\eql(X,x_1) \vee \eql(X,x_2)\)), leading to the expression \((\eql(X,x_1) \vee \eql(X,x_2)) \wedge \eql(Y,y_2)\).
This expression is the result of {\em weakening} literal \(\eql(X,x_2)\) in instance \(\eql(X,x_2) \wedge \eql(Y,y_2)\).
It can be viewed as a candidate explanation of the decision on this instance, just like \(\eql(X,x_2)\),
in the sense that it also represents an abstraction of the instance that preserves the corresponding decision.

For another example, consider the negative decision on instance \(\eql(X,10) \wedge \eql(Y,-20)\), and its corresponding multi-valued expression \(\alpha: \eql(X,x_3) \wedge \eql(Y,y_1)\). Recall that \(x_3\) represents the interval \([6,+\infty)\) and \(y_1\) 
represents the interval \((-\infty,-7)\).
We can drop the characteristic \(\eql(Y,y_1)\) from this instance while guaranteeing that the negative decision will stick (i.e.,
regardless of what value variable \(Y\) takes). Hence, \(\eql(X,x_3)\) is a PI-explanation in this case. But again, if we maintain the
characteristic \(\eql(Y,y_1)\), then this negative decision will stick as long as the value of \(X\) is in the larger, disconnected 
interval \((-\infty,2] \cup [6,+\infty)\). This interval is represented by the expression 
\(\eql(X,x_1) \vee \eql(X,x_3)\) which is a weakening of characteristic \(\eql(X,x_3)\). 
This makes \((\eql(X,x_1) \vee \eql(X,x_3)) \wedge \eql(Y,y_1)\) a candidate explanation as well. 

\subsection{Multi-Valued Literals, Terms and Implicants}

We will now formalize some notions on multi-valued variables and then use them to formally define PI-explanations
in a multi-valued setting~\cite{RameshM94,DBLP:books/daglib/0027780}.
We use three multi-valued variables for our running examples: 
Variable \(A\) with values \(1,2,3\), variable \(B\) with values \(r, b, g\) and  variable \(C\) with values \(l, m, h\).

A {\em literal} is a non-trivial propositional expression that mentions a single variable. 
The following are literals: $\eql(B,r) \vee \eql(B,b)$,  $\eql(A,2)$ and $\neql(C,h)$.
The following are not literals as they are trivial: $\eql(B,r) \vee \eql(B,b) \vee \eql(B,g)$ and $\eql(C,h) \wedge \neql(C,h)$.
Intuitively, for a variable with \(n\) values, a literal specifies a set of values \(S\) where the cardinality of set \(S\) is in \(\{1, \ldots, n-1\}\). 
A literal is {\em simple} if it specifies a single value (cardinality of set \(S\) is \(1\)).
When multi-valued variables correspond to the discretization of continuous variables, our treatment allows a
literal to specify non-contiguous intervals of a continuous variable. 

Consider two literals \(\l_i\) and \(\l_j\) for the same variable.
We say \(\l_i\) is {\em stronger} than \(\l_j\) iff \(\l_i \models \l_j\) and \(\l_i \not \equiv \l_j\). In this case, \(\l_j\) is {\em weaker} than \(\l_i\).
For example,  $\eql(B,r)$ is stronger than $\eql(B,r) \vee \eql(B,b)$.
It is possible to have two literals where neither is stronger or weaker than the other (e.g.,  $\eql(B,r) \vee \eql(B,b)$ and  $\eql(B,g)$).

A {\em term} is a conjunction of literals over distinct variables. The following is a term:
$\eql(A,2) \wedge (\eql(B,r) \vee \eql(B,b)) \wedge \neql(C,h)$.
A term is {\em simple} if all of its literals are simple. 
The following term is simple: 
$\eql(A,2) \wedge \eql(B,r) \wedge \eql(C,h)$.
The following terms are not simple:
$\neql(A,2) \wedge \eql(B,r) \wedge \eql(C,h)$ and
$\eql(A,2) \wedge (\eql(B,r) \vee \eql(B,b)) \wedge \eql(C,h)$.
 A simple term that mentions every variable is called an {\em instance.}

Term \(\t_i\) {\em subsumes} term \(\t_j\) iff \(\t_j \models \t_i\). If we also have
\(\t_i \not \equiv \t_j\), then \(\t_i\) {\em strictly subsumes} \(\t_j\).
For example, the term $\eql(A,2) \wedge (\eql(B,r) \vee \eql(B,b)) \wedge \neql(C,h)$ is strictly subsumed by the terms
$\neql(A,1) \wedge (\eql(B,r) \vee \eql(B,b)) \wedge \neql(C,h)$ and 
$\eql(A,2) \wedge \neql(C,h)$.

We stress two points now. First, if term \(\t_i\) strictly subsumes term \(\t_j\) that
does not necessarily mean that \(\t_i\) mentions a fewer number of variables than \(\t_j\). In fact, it is possible that
the literals of \(\t_i\) and \(\t_j\) are over the same set of variables. Second, a term does not necessarily fix the values of its variables
(unless it is a simple term), which is a departure from how terms are defined in Boolean logic.

In Boolean logic, the only way to get a term that strictly subsumes term \(\t\) is by dropping some literals from \(\t\).
In multi-valued logic, we can also do this by weakening some literals in term \(\t\) (i.e., without dropping any of its variables).
This notion of {\em weakening} a literal generalizes the notion of {\em dropping} a literal in the Boolean setting. 
In particular, dropping a Boolean literal \(\l\) from a Boolean term can be viewed as weakening it into \(\l \vee \neg \l\).

Term \(\t\) is an {\em implicant} of expression \(\Delta\) iff \(\t \models \Delta\).
Term \(\t\) is a {\em prime implicant} of \(\Delta\) iff it is an implicant of \(\Delta\) that is not strictly subsumed 
by another implicant of \(\Delta\). It is possible to have two terms over the same set of variables such that
(a)~the terms are compatible in that they admit some common instance, 
(b)~both are implicants of some expression \(\Delta\), yet 
(c)~only one of them is a prime implicant of \(\Delta\). We stress this possibility as it does not arise in a Boolean setting.
We define the notions of {\em simple implicant} and {\em simple prime implicant} in the expected way.

\subsection{Multi-Valued Explanations}

Consider now a {\em classifier} specified using a multi-valued expression \(\Delta\). The variables of \(\Delta\) will be called {\em features}
so an {\em instance} \(\alpha\) is a simple term that mentions all features. That is, an instance fixes a value
for each feature of the classifier. A decision on instance \(\alpha\) is positive iff the expression \(\Delta\)
evaluates to \(1\) on instance \(\alpha\), written \(\Delta(\alpha)=1\). Otherwise, the decision is negative (when \(\Delta(\alpha)=0\)).

The notation \(\Delta_\alpha\) is crucial for defining explanations: 
\(\Delta_\alpha\) is defined as \(\Delta\) if decision \(\Delta(\alpha)\) is positive and \(\Delta_\alpha\) is defined
as \(\neg \Delta\) if decision \(\Delta(\alpha)\) is negative. 
A {\em PI-explanation} for decision \(\Delta(\alpha)\) is a prime implicant of \(\Delta_\alpha\) that is
consistent with instance \(\alpha\). This basically generalizes the notion of PI-explanation introduced in~\cite{ShihCD18} 
to a multi-valued setting.

The term {\em explanation} is somewhat too encompassing so any definition of this general notion is likely
to draw criticism as being too narrow. The {\em PI-explanation} is indeed narrow as it is based on a  
syntactic restriction: it must be a conjunction of literals (i.e., a term)~\cite{ShihCD18}. In the Boolean
setting, a PI-explanation is a minimal subset of instance characteristics that is sufficient to trigger
the same decision made on the instance. In the multi-valued setting, it can be more generally described
as an {\em abstraction} of the instance that triggers the same decision made on the instance (still in the syntactic form of a term). 

As an example, consider the following truth table representing the decision function of a classifier over two ternary variables \(X\) and \(Y\):
\shrink{
\begin{center}
\small
\begin{tabular}{cc|c}
\(X\) & \(Y\) & \(f(X,Y)\) \\ \hline
\(x_1\) & \(y_1\) & 1 \\ 
\(x_1\) & \(y_2\) & 0 \\ % (X=x1 v X=x2) ^ (Y=y2 v Y=y3)
\(x_1\) & \(y_3\) & 0 \\ % (X=x1 v X=x2) ^ (Y=y2 v Y=y3)
\end{tabular}
\quad
\begin{tabular}{cc|c}
\(X\) & \(Y\) & \(f(X,Y)\) \\ \hline
\(x_2\) & \(y_1\) & 1 \\ 
\(x_2\) & \(y_2\) & 0 \\ % (X=x1 v X=x2) ^ (Y=y2 v Y=y3)
\(x_2\) & \(y_3\) & 0 \\ % (X=x1 v X=x2) ^ (Y=y2 v Y=y3)
\end{tabular}
\quad
\begin{tabular}{cc|c}
\(X\) & \(Y\) & \(f(X,Y)\) \\ \hline
\(x_3\) & \(y_1\) & 1 \\ % X=x3, Y=y1
\(x_3\) & \(y_2\) & 1 \\ % X=x3
\(x_3\) & \(y_3\) & 1 \\ % X=x3
\end{tabular}
\end{center}
}
\begin{center}
\small
\setlength{\tabcolsep}{5pt}
\begin{tabular}{c|c|c|c|c|c|c|c|c|c}
\(X,Y\) & 
\(x_1 y_1\) &
\(x_1 y_2\) &
\(x_1 y_3\) &
\(x_2 y_1\) &
\(x_2 y_2\) &
\(x_2 y_3\) &
\(x_3 y_1\) &
\(x_3 y_2\) &
\(x_3 y_3\) \\ \hline
\(f(X,Y)\) & 1 & 0 & 0 & 1 & 0 & 0 & 1 & 1 & 1
\end{tabular}
\end{center}
Consider instance \(\eql(X,x_3) \wedge \eql(Y,y_1)\) leading to a positive decision.  
The sub-term \(\eql(X,x_3)\) is a PI-explanation for this decision: 
setting input \(X\) to \(x_3\) suffices to trigger a positive decision. 
Similarly, the sub-term \(\eql(Y,y_1)\) is a second PI-explanation for this decision. 
Consider now instance \(\eql(X,x_1) \wedge \eql(Y,y_2)\) leading to a negative decision.  
This decision has a single PI-explanation: \(\neql(X,x_3) \wedge \neql(Y,y_1)\).  
Any instance consistent with this explanation will be decided negatively.

\shrink{
\begin{center}
\begin{tabular}{cc|c}
\(X\) & \(Y\) & \(f(X,Y)\) \\ \hline
\(x_1\) & \(y_1\) & 1 \\ 
\(x_1\) & \(y_2\) & 0 \\ % (X=x1 v X=x2) ^ (Y=y2 v Y=y3)
\(x_1\) & \(y_3\) & 0 \\ % (X=x1 v X=x2) ^ (Y=y2 v Y=y3)
\(x_2\) & \(y_1\) & 1 \\ 
\(x_2\) & \(y_2\) & 0 \\ % (X=x1 v X=x2) ^ (Y=y2 v Y=y3)
\(x_2\) & \(y_3\) & 0 \\ % (X=x1 v X=x2) ^ (Y=y2 v Y=y3)
\(x_3\) & \(y_1\) & 1 \\ % X=x3, Y=y1
\(x_3\) & \(y_2\) & 1 \\ % X=x3
\(x_3\) & \(y_3\) & 1 \\ % X=x3
\end{tabular}
\end{center}
}

\section{Encoding Multi-Valued Behavior}
\label{sec:encodings}

We next discuss three encodings that we tried for the purpose of symbolically representing the behavior of decision trees (and 
random forests). The first two encodings turned out unsuitable for computing prime implicants. Here, {\em suitability} 
refers to the ability of computing multi-valued prime implicants by processing Boolean prime implicants {\em locally} and {\em independently.}
The third encoding, based on a classical encoding~\cite{DBLP:conf/ijcai/Kleer89}, was 
suitable for this purpose but required a usage that deviates from tradition. Using this encoding in a classical way makes it unsuitable as well.
The summary of the findings below is that while an encoding may be appropriate for testing satisfiability or counting models,
it may not be suitable for computing prime implicants (and, hence, explanations). While much attention was given to encodings in
the context of satisfiability and model counting, we are not aware of similar treatments for computing prime implicants. 

\subsection{Prefix Encoding}

Consider a multi-valued variable \(X\) with values \(x_1, \ldots, x_{n}\).
This encoding uses Boolean variables \(x_2, \ldots, x_{n}\) to encode the values of variable \(X\).
Literal \(\eql(X,x_i)\) is encoded by setting the first \(i-1\) Boolean variables to \(1\) and the rest to \(0\). 
For example, if \(n=3\), the values of \(X\) are encoded as \(\n{x}_2 \n{x}_3\), 
 \(x_2 \n{x}_3\) and  \(x_2 x_3\).
Some instantiations of these Boolean variables will not correspond to any
value of variable \(X\) and are ruled out by enforcing the following constraint:
all Boolean variables set to \(1\) must occur before all Boolean variables set to \(0\). 
We denote this constraint by \(\Psi_X\): \(\bigwedge_{i \in \{3,\ldots,n\}} (x_{i} \Rightarrow x_{i-1})\).

The fundamental problem with this encoding is that a multi-valued literal that represents non-contiguous 
values cannot be represented by a Boolean term. Hence, this encoding cannot generate prime implicants that include such literals. Consider the multi-valued expression \(\Delta = (\eql(X,x_1) \vee \eql(X,x_3))\),
where \(X\) has values \(x_1, \ldots, x_4\), 
and its Boolean encoding \(\Delta_b = \n{x}_2 \n{x}_3 \n{x}_4 + x_2 x_3 \n{x}_4\). 
There is only one prime implicant of \(\Delta\), which is \(\eql(X,x_1) \vee \eql(X,x_3)\), but this prime
implicant cannot be represented by a Boolean term (that implies \(\Delta_b\)) so it will never be generated.

\subsection{Highest-Bit Encoding}

Consider a multi-valued variable \(X\) with values \(x_1,x_2, \ldots, x_n\).
This encoding uses Boolean variables \(x_2, x_3, \ldots, x_{n}\) to encode the values of variable \(X\).
Every instantiation of these Boolean variables will map to a value of variable \(X\) in the following way. 
If all Boolean variables are \(0\), then we map the instantiation to value \(x_1\). 
Otherwise we map an instantiation to the maximum index whose variable is \(1\).
The following table provides an example for \(n=4\).
\shrink{
\begin{center}
\small
\begin{tabular}{c|c|c}
  \(x_2 x_3 x_4\)     & $1$-index  & value  \\ \hline
  000          & - & \(x_1\) \\
  001          & 4 & \(x_4\) \\
  010          & 3 & \(x_3\) \\
  011          & 4 & \(x_4\) \\
\end{tabular}
\quad
\begin{tabular}{c|c|c}
  \(x_2 x_3 x_4\)     & $1$-index  & value  \\ \hline
  100          & 2 & \(x_2\) \\
  101          & 4 & \(x_4\) \\
  110          & 3 & \(x_3\) \\
  111          & 4 & \(x_4\) \\
\end{tabular}
\end{center}
}
\begin{center}
\small
\setlength{\tabcolsep}{5pt}
\begin{tabular}{c|c|c|c|c|c|c|c|c}
\(x_2 x_3 x_4\) & 
000 &
001 &
010 &
011 &
100 &
101 &
110 &
111 \\ \hline
highest $1$-index       & 
- &
4 &
3 &
4 &
2 &
4 &
3 &
4 \\ \hline
value &
\(x_1\) &
\(x_4\) &
\(x_3\) &
\(x_4\) &
\(x_2\) &
\(x_4\) &
\(x_3\) &
\(x_4\) 
\end{tabular}
\end{center}
We can alternatively view this encoding as representing 
literal \(\eql(X,x_1)\) using the Boolean term \(\n{x}_2\ldots\n{x}_n\) and
literal \(\eql(X,x_i)\), \(i \geq 2\), using the term  \(x_i\n{x}_{i+1}\ldots\n{x}_n\).
Literals over multiple values can also be represented with this encoding.  
For example, we can represent the literal \(\eql(X,x_1) \vee \eql(X,x_2)\) using the term \(\n{x}_3\n{x}_4.\)
\shrink{
The following table provides some encodings for a variable \(X\) with four values:
{\small
\renewcommand{\arraystretch}{1.15}
\[
\begin{array}{r|c|l}
\mbox{Expression} & \mbox{Encoding} & \mbox{Boolean Terms for Values} \\ \hline
\eql(X,x_1) & \n{x}_2\n{x}_3\n{x}_4 & \n{x}_2\n{x}_3\n{x}_4 \\
\eql(X,x_2) & x_2\n{x}_3\n{x}_4 & x_2\n{x}_3\n{x}_4 \\
\eql(X,x_3) & x_3\n{x}_4 & \n{x}_2 x_3\n{x}_4 + x_2 x_3\n{x}_4 \\
\eql(X,x_4) & x_4 & \n{x}_2 \n{x}_3 x_4 + \n{x}_2 x_3 x_4 + x_2 \n{x}_3 x_4 + x_2 x_3 x_4 \\ 
\eql(X,x_1) \vee \eql(X,x_2) & \n{x}_3\n{x}_4 & \n{x}_2\n{x}_3\n{x}_4 + x_2\n{x}_3\n{x}_4 \\ 
\eql(X,x_1) \vee \eql(X,x_3) & \n{x}_2\n{x}_4 + x_3\n{x}_4 & \n{x}_2\n{x}_3\n{x}_4 + x_3\n{x}_4 \\ 
\eql(X,x_1) \vee \eql(X,x_2) \vee \eql(X,x_3) & \n{x}_4 & \n{x}_2\n{x}_3\n{x}_4 + x_2\n{x}_3\n{x}_4 + x_3\n{x}_4 \\ 
\eql(X,x_1) \vee \eql(X,x_2) \vee \eql(X,x_3) \vee \eql(X,x_4) & \top & \n{x}_2\n{x}_3\n{x}_4 + x_2\n{x}_3\n{x}_4 + x_3\n{x}_4 + x_4
\end{array}
\]
\renewcommand{\arraystretch}{1}
}
}

This encoding also turned out to be unsuitable for computing prime implicants.
Consider the multi-valued expression \(\Delta = (\eql(X,x_1) \vee \eql(X,x_3))\), which has one prime implicant \(\Delta\).
The Boolean encoding \(\Delta_b\) is \(\n{x}_2\n{x}_3\n{x}_4 + x_3\n{x}_4\) and has \emph{two} prime implicants
\(\n{x}_2\n{x}_4\) and \(x_3\n{x}_4\). The term \(x_3\n{x}_4\) corresponds to the multi-valued implicant \(\eql(X,x_3)\), which is not prime.  
The term \(\n{x}_2\n{x}_4\) does not even correspond to a multi-valued term.
So in this encoding too,  prime implicants of the original multi-valued expression \(\Delta\)
cannot be computed by locally and independently processing prime implicants of the encoded Boolean expression \(\Delta_b\).

\subsection{One-Hot Encoding}

The prefix and highest-bit encodings provide some insights into requirements that enable one to locally and independently
map Boolean prime implicants into multi-valued ones. The requirements are:
%\begin{enumerate}
(1) every multi-valued literal should be representable using a Boolean term, and 
(2) equivalence and subsumption relations over multi-valued literals should be preserved over
their Boolean encodings. 
%\end{enumerate}
The next encoding satisfies these requirements. It is based on~\cite{DBLP:conf/ijcai/Kleer89} but
deviates from it in some significant ways that we explain later. 

Suppose \(X\) is a multi-valued variable with values \(x_1, \ldots, x_n\). This encoding uses a Boolean variable
\(x_i\) for each value \(x_i\) of variable \(X\). Suppose now that \(\l\) is a literal that specifies a subset \(S\) of these
values. The literal will be encoded using the {\em negative} Boolean term \(\bigwedge_{x_i \not \in S} \n{x}_i\). 
For example, if variable \(X\) has three values, then literal \(\eql(X,x_2)\) will be encoded using the negative Boolean 
term \(\n{x}_1\n{x}_3\) and literal \(\eql(X,x_1) \vee \eql(X,x_2)\) will be encoded using the negative Boolean 
term \(\n{x}_3\). This encoding requires the employment of an exactly-one constraint for each variable \(X\),
which we denote by \(\Psi_X\): \((\bigvee_{i} x_i) \wedge \bigwedge_{i \neq j} \neg (x_i \wedge x_j)\).
We also use \(\Psi\) to denote the conjunction of all exactly-one constraints.

Using the encoding in~\cite{DBLP:conf/ijcai/Kleer89}, one typically represents literal \(\eql(X,x_i)\) by the Boolean term \(x_i\)
which {\em asserts} value \(x_i\). Our encoding, however, represents this literal by {\em eliminating} all
other values of \(X\). The following result reveals why we made this choice (proofs of results can be found in the appendix). 

\begin{proposition}\label{prop:oh}
Multi-valued terms correspond one-to-one to negative Boolean terms
that are consistent with  \(\Psi\).
Equivalence and subsumption relations on multi-valued terms are preserved on their Boolean encodings.
\end{proposition}

Exactly-one constraints are normally added to an encoding as done in~\cite{DBLP:conf/ijcai/Kleer89}. We next
show that this leads to unintended results when computing prime implicants, requiring another deviation from~\cite{DBLP:conf/ijcai/Kleer89}.
Consider two ternary variables \(X\) and \(Y\), the
expression \(\Delta: \eql(X,x_1) \vee \eql(Y,y_1)\) and its Boolean encoding \(\Delta_b:\n{x}_2\n{x}_3 + \n{y}_2 \n{y}_3\).
If \(\Psi\) is the conjunction of all exactly-one constraints (\(\Psi=\Psi_X \wedge \Psi_Y\)), 
then \(\Delta\) and \(\Delta_b \wedge \Psi\) will each have five models:
\shrink{
\begin{center}
\renewcommand{\arraystretch}{1.25}
\begin{tabular}{c|c}
\(\Delta\) & \(\Delta_b \wedge \Psi\) \\ \hline
\(\eql(X,x_1), \eql(Y,y_1)\) & \(x_1 \n{x}_2 \n{x}_3 y_1 \n{y}_2 \n{y}_3\) \\
\(\eql(X,x_1), \eql(Y,y_2)\) & \(x_1 \n{x}_2 \n{x}_3 \n{y}_1 y_2 \n{y}_3\) \\
\(\eql(X,x_1), \eql(Y,y_3)\) & \(x_1 \n{x}_2 \n{x}_3 \n{y}_1 \n{y}_2 y_3\) \\
\(\eql(X,x_2), \eql(Y,y_1)\) & \(\n{x}_1 x_2 \n{x}_3 y_1 \n{y}_2 \n{y}_3\) \\
\(\eql(X,x_3), \eql(Y,y_1)\) & \(\n{x}_1 \n{x}_2 x_3 y_1 \n{y}_2 \n{y}_3\) \\
\end{tabular}
\end{center}
}
\begin{center}
\small
\setlength{\tabcolsep}{3.5pt}
\renewcommand{\arraystretch}{1.25}
\begin{tabular}{c|c|c|c|c|c}
\(\Delta\) & 
\(\eql(X,x_1), \eql(Y,y_1)\) & 
\(\eql(X,x_1), \eql(Y,y_2)\) & 
\(\eql(X,x_1), \eql(Y,y_3)\) & 
\(\eql(X,x_2), \eql(Y,y_1)\) & 
\(\eql(X,x_3), \eql(Y,y_1)\) \\\hline
\(\Delta_b \wedge \Psi\) &
\(x_1 \n{x}_2 \n{x}_3 y_1 \n{y}_2 \n{y}_3\) & 
\(x_1 \n{x}_2 \n{x}_3 \n{y}_1 y_2 \n{y}_3\) & 
\(x_1 \n{x}_2 \n{x}_3 \n{y}_1 \n{y}_2 y_3\) & 
\(\n{x}_1 x_2 \n{x}_3 y_1 \n{y}_2 \n{y}_3\) & 
\(\n{x}_1 \n{x}_2 x_3 y_1 \n{y}_2 \n{y}_3\) 
\end{tabular}
\end{center}
The term \(\eql(X,x_1)\) is an implicant of \(\Delta\).  
However, its corresponding Boolean encoding \(\n{x}_2 \n{x}_3\) is not an implicant of \(\Delta_b \wedge \Psi\)
(neither is \(x_1\n{x}_2 \n{x}_3\)).
For example, \(x_1 \n{x}_2 \n{x}_3 y_1 y_2 \n{y}_3\) does not imply \(\Delta_b \wedge \Psi\)
since \(y_1 y_2 \n{y}_3\) does not satisfy the exactly-one constraint \(\Psi_Y\).
This motivates Definition~\ref{def:models} below and further results on handling exactly-one constraints, 
which we introduce after some notational conventions.

In what follows, we use \(\Delta\)/\(\t\) to denote multi-valued expressions/terms,
and \(\Gamma\)/\(\r\) to denote Boolean expressions/terms.
We also use \(\Delta_b\) and \(\t_b\) to denote the Boolean encodings of \(\Delta\) and \(\t\).
A {\em completion} of a term is a complete variable instantiation that is consistent with the term.
We use \(\alpha\) to denote completions.
Finally, we use \(\Psi\) to denote the conjunction of all exactly-one constraints.

\begin{definition} \label{def:models}
We define \(\r \models_{\Psi} \Gamma\) iff  \(\alpha \models \Gamma\) for all completions \(\alpha\)
of Boolean term \(\r\) that are consistent with constraint \(\Psi\).
\end{definition}
Note that \(\r \models \Gamma\) implies \(\r \models_{\Psi} \Gamma\) but the converse is not true.
 
\begin{proposition} \label{prop:models}
\(\r \models_{\Psi} \Gamma\) iff \(\r \models (\Psi \Rightarrow \Gamma).\)
\end{proposition}

We now show how one-hot encodings can be used for computing prime implicants, particularly, how exactly-one constraints should be integrated.
\begin{proposition} \label{prop:implicant}
If \(\t\) is a term, then \(\t \models \Delta\) iff \(\t_b \models (\Psi \Rightarrow \Delta_b).\)
\end{proposition}

The proof is based on two lemmas that hold by construction and that use the notion of {\em full encoding} of an instance.
Consider ternary variables \(X\) and \(Y\). For instance \(\t: \eql(X,x_1)\wedge\eql(Y,y_1)\) the full encoding 
is \(\r: x_1\n{x}_2\n{x}_3 y_1\n{y}_2 \n{y}_3\) (\(x_1\) and \(y_1\) are included). 
Note that \(\r \wedge \Psi = \r\) since \(\r\) is guaranteed to satisfy constraints \(\Psi\).

\begin{lemma} \label{lemma:encoding}
If \(\t\) is an instance and \(\r\) is its full encoding, then \(\t \models \Delta\) iff \(\r \models \Delta_b\).
\end{lemma}

\begin{lemma} \label{lemma:completion}
For term \(\t\), there is a one-to-one correspondence between the completions of \(\t\)
and the completions of \(\t_b\) that are consistent with \(\Psi\).
\end{lemma}
Term \(\t: \eql(X,x_1)\vee\eql(X,x_2)\) has six completions: 
\(\eql(X,x_1)\wedge\eql(Y,y_1)\), \(\eql(X,x_2)\wedge\eql(Y,y_1)\), \ldots, \(\eql(X,x_2)\wedge\eql(Y,y_3)\).
Its Boolean encoding \(\t_b: \n{x}_3\) also has six completions that are consistent with \(\Psi\):
 \(x_1\n{x}_2\n{x}_3y_1\n{y}_2\n{y_3}\), \(\n{x}_1 x_2\n{x}_3y_1\n{y}_2\n{y_3}\), \ldots, \(\n{x}_1 x_2\n{x}_3\n{y}_1\n{y}_2 y_3\).
 Each of these completions \(\alpha\) is guaranteed to satisfy constraints \(\Psi\) leading to \(\alpha \wedge \Psi = \alpha\).
Next, we relate the prime implicants of multi-valued expressions and their encodings.

\begin{proposition}\label{prop:pi}
Consider a multi-valued expression \(\Delta\) and its Boolean encoding \(\Delta_b\).
If \(\t\) is a prime implicant of \(\Delta\), then \(\t_b\) is a negative term, consistent with \(\Psi\) and a prime implicant of \(\Psi \Rightarrow \Delta_b\).
If \(\r\) is a prime implicant of \(\Psi \Rightarrow \Delta_b\), negative and  consistent with \(\Psi\), then \(\r\) encodes a  prime implicant of \(\Delta\).
\end{proposition}

This proposition suggests the following procedure for computing multi-valued prime implicants from Boolean
prime implicants. Given a multi-valued expression \(\Delta\), we encode each literal in \(\Delta\)
using its negative Boolean term, leading to the Boolean expression \(\Delta_b\). We then
construct the exactly-one constraints \(\Psi\) and compute prime implicants of \(\Psi \Rightarrow \Delta_b\),
keeping those that are negative and consistent with constraints \(\Psi\).\footnote{It is straightforward to augment the algorithm of \cite{ShihCD18} so that it only enumerates such prime implicants, by blocking the appropriate branches.} Those Boolean prime implicants
correspond precisely to the multi-valued prime implicants of \(\Delta\).\footnote{Note that when computing PI-explanations,
we are interested only in prime implicants that are consistent with a given instance. Any negative prime
implicant which is consistent with an instance must also be consistent with constraints \(\Psi\).  The
only way a negative Boolean term \(\r\) can violate constraints \(\Psi\) is by setting all Boolean variables of
some multi-valued variable to false. However, every instance \(\alpha\) will set one of these Boolean variables
to true so \(\r\) cannot be consistent with \(\alpha\).}

The only system we are aware of that computes prime implicants of decision tree encodings (and forests) is
Xplainer~\cite{JoaoApp}.
This system bypasses the encoding complications we alluded to earlier as it computes prime 
implicants in a specific manner~\cite{IgnatievMM16,IgnatievNM19a}. In particular, it encodes a multi-valued 
expression into a Boolean expression using the classical one-hot encoding. 
But rather than computing prime implicants of the Boolean encoding directly (which would lead to 
incorrect results), it  reduces the problem of computing prime implicants of a multi-valued 
expression into one that requires only consistency testing of the Boolean encoding, which can be done 
using repeated calls to a SAT solver. The classical one-hot encoding is sound and complete for this purpose.
Our treatment, however, is meant to be independent of the specific algorithm used to compute
prime implicants. It would be needed, for example, when compiling the encoding into a tractable
circuit and then computing prime implicants as done in~\cite{ShihCD18,DarwicheHirth20a}.

\subsection{Encoding Decision Trees and Random Forests}

Consider a decision tree, such as the one depicted in Figure~\ref{fig:dt}.  Each internal node in the tree represents a decision, which is either true or false.  Each leaf is annotated with the predicted label.  We can thus view a decision tree as a function whose inputs are all of the unique decisions that can be made in the tree, and whose output is the resulting label.  Each leaf of the decision tree represents a simple term over the decisions made on the path to reach it, found by conjoining the appropriate literals.  The Boolean function representing a particular class can then be found by simply disjoining the paths for all leaves of that class.  That is, this Boolean function outputs true for all inputs that result in the corresponding class label, and false otherwise.  We can also obtain this function for an ensemble of decision trees, such as a random forest.  We first obtain the Boolean functions of each individual decision tree, and then aggregate them appropriately.  For a random forest, we can use a simple majority gate whose inputs are the outputs of each decision tree; see also~\cite{Audemardetal20}.  Finally, once we have the Boolean function of a classifier, we could apply a SAT or SMT solver to analyze it as proposed by \cite{KatzBDJK17,NarodytskaKRSW18,IgnatievNM19a}.  We could also compile it into a tractable representation, such as an Ordered Binary Decision Diagram (OBDD), and then analyze it as proposed by \cite{ShihCD18,ShihCD18b,SDC19,shi2020tractable}.  In the latter case, a representation such as an OBDD allows us to perform certain 
queries and transformation on a Boolean function efficiently, which facilitates the explanation and formal verification of the underlying machine 
learning classifier, as also shown more generally in~\cite{Audemardetal20}.

\section{Conclusion}
\label{sec:conclusion}

We considered the encoding of input-output behavior of decision trees and random forests
using Boolean expressions. Our focus has been on the
suitability of encodings for computing prime implicants, which have
recently played a central role in explaining the decisions of machine learning classifiers.
Our findings have identified a particular encoding that is suitable for this purpose. Our encoding is based on a classical
encoding that has been employed for the task of satisfiability but that can lead to incorrect
results when computing prime implicants, which further emphasizes the merit of the
investigation we conducted in this paper. 

\vspace{2mm}

\noindent{\bf Ack.}
This work has been partially supported by grants from NSF IIS-1910317, 
ONR N00014-18-1-2561, DARPA N66001-17-2-4032 and a gift from JP Morgan.

\bibliographystyle{splncs04}
\bibliography{bib/kr20,bib/ecai20}

\begin{thebibliography}{10}
\providecommand{\url}[1]{\texttt{#1}}
\providecommand{\urlprefix}{URL }
\providecommand{\doi}[1]{https://doi.org/#1}

\bibitem{Audemardetal20}
Audemard, G., Koriche, F., Marquis, P.: On tractable {XAI} queries based on
  compiled representations. In: Proc. of KR'20 (2020), to appear

\bibitem{DBLP:conf/cp/BessiereHO09}
Bessiere, C., Hebrard, E., O'Sullivan, B.: Minimising decision tree size as
  combinatorial optimisation. In: {CP}. Lecture Notes in Computer Science,
  vol.~5732, pp. 173--187. Springer (2009)

\bibitem{ChoiShiShihDarwiche18}
Choi, A., Shi, W., Shih, A., Darwiche, A.: Compiling neural networks into
  tractable {B}oolean circuits. In: AAAI Spring Symposium on Verification of
  Neural Networks (VNN) (2019)

\bibitem{BooleanFunctions}
Crama, Y., Hammer, P.L.: Boolean Functions - Theory, Algorithms, and
  Applications, Encyclopedia of mathematics and its applications, vol.~142.
  Cambridge University Press (2011)

\bibitem{DarwicheHirth20a}
Darwiche, A., Hirth, A.: On the reasons behind decisions. In: Proceedings of
  the 24th European Conference on Artificial Intelligence (ECAI) (2020)

\bibitem{IgnatievMM16}
Ignatiev, A., Morgado, A., Marques{-}Silva, J.: Propositional abduction with
  implicit hitting sets. In: Proceedings of the 22nd European Conference on
  Artificial Intelligence (ECAI). pp. 1327--1335 (2016)

\bibitem{IgnatievNM19a}
Ignatiev, A., Narodytska, N., Marques{-}Silva, J.: Abduction-based explanations
  for machine learning models. In: Proceedings of the Thirty-Third Conference
  on Artificial Intelligence ({AAAI}). pp. 1511--1519 (2019)

\bibitem{IgnatievNM19b}
Ignatiev, A., Narodytska, N., Marques{-}Silva, J.: On relating explanations and
  adversarial examples. In: Advances in Neural Information Processing Systems
  32 (NeurIPS). pp. 15857--15867 (2019)

\bibitem{JoaoApp}
Ignatiev, A., Narodytska, N., Marques{-}Silva, J.: On validating, repairing and
  refining heuristic {ML} explanations. CoRR  \textbf{abs/1907.02509} (2019)

\bibitem{KatzBDJK17}
Katz, G., Barrett, C.W., Dill, D.L., Julian, K., Kochenderfer, M.J.: Reluplex:
  An efficient {SMT} solver for verifying deep neural networks. In: Computer
  Aided Verification {CAV}. pp. 97--117 (2017)

\bibitem{DBLP:conf/ijcai/Kleer89}
de~Kleer, J.: A comparison of {ATMS} and {CSP} techniques. In: {IJCAI}. pp.
  290--296. Morgan Kaufmann (1989)

\bibitem{Leofante18}
Leofante, F., Narodytska, N., Pulina, L., Tacchella, A.: Automated verification
  of neural networks: Advances, challenges and perspectives. CoRR
  \textbf{abs/1805.09938} (2018)

\bibitem{mccluskey}
{McCluskey}, E.J.: Minimization of boolean functions. The Bell System Technical
  Journal  \textbf{35}(6),  1417--1444 (Nov 1956)

\bibitem{DBLP:books/daglib/0027780}
Miller, D.M., Thornton, M.A.: Multiple Valued Logic: Concepts and
  Representations, Synthesis lectures on digital circuits and systems, vol.~12.
  Morgan {\&} Claypool Publishers (2008)

\bibitem{NarodytskaIPM18}
Narodytska, N., Ignatiev, A., Pereira, F., Marques{-}Silva, J.: Learning
  optimal decision trees with {SAT}. In: Lang, J. (ed.) Proceedings of the
  Twenty-Seventh International Joint Conference on Artificial Intelligence
  ({IJCAI}). pp. 1362--1368 (2018)

\bibitem{NarodytskaKRSW18}
Narodytska, N., Kasiviswanathan, S.P., Ryzhyk, L., Sagiv, M., Walsh, T.:
  Verifying properties of binarized deep neural networks. In: Proceedings of
  the Thirty-Second AAAI Conference on Artificial Intelligence (AAAI) (2018)

\bibitem{quine1}
Quine, W.V.: The problem of simplifying truth functions. The American
  Mathematical Monthly  \textbf{59}(8),  521--531 (1952)

\bibitem{quine2}
Quine, W.V.: On cores and prime implicants of truth functions. The American
  Mathematical Monthly  \textbf{66}(9),  755--760 (1959)

\bibitem{RameshM94}
Ramesh, A., Murray, N.V.: Computing prime implicants/implicates for regular
  logics. In: Proceedings of the 24th {IEEE} International Symposium on
  Multiple-Valued Logic ({ISMVL}). pp. 115--123 (1994)

\bibitem{Renooij18}
Renooij, S.: Same-decision probability: Threshold robustness and application to
  explanation. In: Studeny, M., Kratochvil, V. (eds.) Proceedings of the
  International Conference on Probabilistic Graphical Models ({PGM}).
  Proceedings of Machine Learning Research, vol.~72, pp. 368--379. {PMLR}
  (2018)

\bibitem{anchors:aaai18}
Ribeiro, M.T., Singh, S., Guestrin, C.: Anchors: High-precision model-agnostic
  explanations. In: Proceedings of the Thirty-Second AAAI Conference on
  Artificial Intelligence (AAAI) (2018)

\bibitem{ANCHOR}
Ribeiro, M.T., Singh, S., Guestrin, C.: Anchors: High-precision model-agnostic
  explanations. In: {AAAI}. pp. 1527--1535. {AAAI} Press (2018)

\bibitem{shi2020tractable}
Shi, W., Shih, A., Darwiche, A., Choi, A.: On tractable representations of
  binary neural networks. In: Proc. of KR'20 (2020), to appear

\bibitem{ShihCD18b}
Shih, A., Choi, A., Darwiche, A.: Formal verification of bayesian network
  classifiers. In: {PGM}. Proceedings of Machine Learning Research, vol.~72,
  pp. 427--438. {PMLR} (2018)

\bibitem{ShihCD18}
Shih, A., Choi, A., Darwiche, A.: A symbolic approach to explaining bayesian
  network classifiers. In: {IJCAI}. pp. 5103--5111. ijcai.org (2018)

\bibitem{SDC19}
Shih, A., Darwiche, A., Choi, A.: Verifying binarized neural networks by
  angluin-style learning. In: SAT (2019)

\bibitem{DBLP:conf/cp/Walsh00}
Walsh, T.: {SAT} v {CSP}. In: {CP}. Lecture Notes in Computer Science,
  vol.~1894, pp. 441--456. Springer (2000)

\end{thebibliography}

\appendix

\section{Proofs}

\begin{proof}[of Proposition~\ref{prop:oh}]
For multi-valued term \(\t\), the Boolean encoding \(\t_b\) is a negative term and consistent with \(\Psi\) by construction.
Suppose now that \(\r\) is a negative Boolean term that is consistent with \(\Psi\). If \(\r\) mentions a Boolean variable
of multi-valued variable \(X\), then \(\r\) cannot mention all Boolean variables of \(X\), otherwise \(\r\) will be ruling
out all possible values of \(X\) and hence inconsistent with \(\Psi\). Hence, \(\r\) encodes a literal over variable \(X\)
when \(\r\) mentions a Boolean variable for \(X\). More generally, \(\r\) encodes a term over multi-valued variables whose Boolean variables are
mentioned in \(\r\). To prove the second part of the theorem, consider literals \(\l_1\) and \(\l_2\),
which specify values \(S_1\) and \(S_2\) for variable \(X\). The two literals are equivalent iff \(S_1=S_2\) iff
\(\bigwedge_{x_i \not \in S_1} \n{x}_i\) and  \(\bigwedge_{x_i \not \in S_2} \n{x}_i\) are equivalent. 
Moreover, \(\l_1 \models \l_2\) iff \(S_1 \subseteq S_2\) iff
\(\bigwedge_{x_i \not \in S_1} \n{x}_i \models \bigwedge_{x_i \not \in S_2} \n{x}_i\).
%For example, \(\eql(X,x_1)\vee\eql(X,x_2) \models \eql(X,x_1)\vee\eql(X,x_2)\vee\eql(X,x_4)\) so \(\n{x}_3\n{x}_4 \models \n{x}_3\). 
Equivalence and subsumption relations are then preserved on literals, and on terms as well. 
%For example, \(\eql(X,x_1)\wedge \eql(Y,y_1) \models \eql(X,x_1)\vee\eql(X,x_4)\) 
%so \(\n{x}_2\n{x}_3\n{x}_4 \n{y}_2\n{y}_3\n{y}_4 \models \n{x}_2\n{x}_3\).
\end{proof}

\begin{proof}[of Proposition~\ref{prop:models}]
(\(\Rightarrow\)) 
Suppose \(\r \models_\Psi \Gamma\) and let \(\c\) be a completion of \(\r\).
If \(\c\) is consistent with \(\Psi\), then \(\c \models \Gamma\) by Definition~\ref{def:models}. 
If \(\c\) is not consistent with \(\Psi\), then \(\c \models \neg \Psi\). Hence, \(\r \models \neg \Psi \vee \Gamma.\)
(\(\Leftarrow\)) 
Suppose \(\r \models \neg \Psi \vee \Gamma\) and let \(\c\) be a completion of \(\r\) that is consistent with \(\Psi\).
Then \(\c \models \neg \Psi \vee \Gamma\) and, hence,  \(\c \wedge \Psi \models \Gamma\)
and  \(\c \models \Gamma\). We then have \(\r \models_{\Psi} \Gamma\) by Definition~\ref{def:models}. 
\end{proof}

\begin{proof}[of Proposition~\ref{prop:implicant}]
(\(\Rightarrow\)) Suppose \(\tau \models \Delta.\) Then \(\alpha \models \Delta\) for all completions \(\alpha\) of \(\t\).   
By Lemmas~\ref{lemma:encoding} and~\ref{lemma:completion}, \(\alpha_b \models \Delta_b\) for all
completions \(\alpha_b\) of \(\t_b\) that are consistent with \(\Psi\). Hence \(\tau_b \models \neg \Psi \vee \Delta_b.\)
(\(\Leftarrow\)) Suppose \(\tau_b \models \neg \Psi \vee \Delta_b\) and let \(\alpha_b\) be 
a completion of \(\t_b\) (\(\alpha_b \models \neg \Psi \vee \Delta_b\)).
For each \(\alpha_b\) consistent with \(\Psi\), we have \(\alpha_b \models \Psi\) and hence \(\alpha_b \models \Delta_b.\)
By Lemmas~\ref{lemma:encoding} and~\ref{lemma:completion}, the completions \(\alpha\) of \(\t\) correspond
to these \(\alpha_b\) (consistent with \(\Psi\)), leading to \(\alpha \models \Delta\) and hence \(\t \models \Delta\).
\end{proof}

\begin{proof}[of Proposition~\ref{prop:pi}]
(\(\Rightarrow\))
Suppose \(\t\) is a prime implicant of \(\Delta\). Then \(\t \models \Delta\). Moreover, \(\t_b\ \models (\Psi \Rightarrow \Delta_b)\)
by Proposition~\ref{prop:implicant} so \(\t_b\) is an implicant of \(\Psi \Rightarrow \Delta_b\) (\(\t_b\) is negative and consistent
with \(\Psi\) by construction).
Suppose \(\t_b\) is not a prime implicant of \(\Psi \Rightarrow \Delta_b\). Then \(\r \models (\Psi \Rightarrow \Delta_b)\) for
a strict subset \(\r\) of \(\t_b\), which must be consistent with \(\Psi\)  since \(\t_b \supset \r\) is consistent with \(\Psi\).
Hence, \(\r\) encodes a term \(\t^\star\) that is strictly weaker than term \(\t\) by Proposition~\ref{prop:oh}. 
Moreover, \(\t^\star \models \Delta\) by Proposition~\ref{prop:implicant} so \(\t\) is not a prime implicant of \(\Delta\), which is a contradiction.
Therefore, \(\t_b\) is a prime implicant of \(\Psi \Rightarrow \Delta_b\).
(\(\Leftarrow\)) 
Suppose \(\r\) is a prime implicant of \(\Psi \Rightarrow \Delta_b\), negative and consistent with \(\Psi\). Then
\(\r\) encodes a term \(\t\) by Proposition~\ref{prop:oh}.
Moreover,  \(\r = \t_b \models \Psi \Rightarrow \Delta_b\) so \(\t \models \Delta\) by Proposition~\ref{prop:implicant}.
Hence, \(\t\) is an implicant of \(\Delta\).
Suppose now that \(\t^\star \models \Delta\) for some term \(\t^\star\) that is strictly weaker than term \(\t\). 
Then \(\t^\star_b \models \Psi \Rightarrow \Delta_b\) by Proposition~\ref{prop:implicant}. This means \(\r\) is not a prime
implicant of \(\Psi \Rightarrow \Delta_b\) since  \(\t^\star_b \subset \t_b = \r\) by Proposition~\ref{prop:oh}, 
which is a contradiction. Hence, the term \(\t\) encoded by \(\r\) is a prime implicant of \(\Delta\).
\end{proof}

\end{document}